\documentclass[12pt]{article}
\usepackage{amsmath}
\usepackage{times}
\usepackage{graphicx}
\usepackage{color}
\usepackage{multirow}
\usepackage[authoryear]{natbib}
\usepackage{rotating}
\usepackage{bbm}
\usepackage{latexsym}
\usepackage{setspace}

\newcommand{\captionfonts}{\normalsize}

\makeatletter  
\long\def\@makecaption#1#2{%
  \vskip\abovecaptionskip
  \sbox\@tempboxa{{\captionfonts #1: #2}}%
  \ifdim \wd\@tempboxa >\hsize
    {\captionfonts #1: #2\par}
  \else
    \hbox to\hsize{\hfil\box\@tempboxa\hfil}%
  \fi
  \vskip\belowcaptionskip}
\makeatother   

\newcommand{\sket}[1]{{\ensuremath{\lvert#1\rangle}}}
\newcommand{\lket}[1]{{\ensuremath{\left\lvert#1\right\rangle}}}
\newcommand{\tr}[1]{{\text{Tr}}}
\newcommand{\ket}[1]{\if@display\lket{#1}\else\sket{#1}\fi}
\newcommand{\sbra}[1]{{\ensuremath{\langle#1\rvert}}}
\newcommand{\lbra}[1]{{\ensuremath{\left\langle#1\right\rvert}}}
\newcommand{\bra}[1]{\if@display\lbra{#1}\else\sbra{#1}\fi}
\newcommand{\sbraket}[2]{{\ensuremath{\langle#1\rvert#2\rangle}}}
\newcommand{\lbraket}[2]{{\ensuremath{\left\langle#1\!\left\rvert\vphantom{#1}#2\right.\!\right\rangle}}}
\newcommand{\braket}[2]{\if@display\lbraket{#1}{#2}\else\sbraket{#1}{#2}\fi}

\newcommand{\sketbra}[2]{{\ensuremath{\lvert #1\rangle\!\langle #2\rvert}}}
\newcommand{\lketbra}[2]{{\ensuremath{\left\lvert #1\right\rangle\!\!\left\langle #2\right\rvert}}}
\newcommand{\ketbra}[2]{\if@display\lketbra{#1}{#2}\else\sketbra{#1}{#2}\fi}

\begin{document}
 
\hspace{13.9cm}1

\ \vspace{20mm}\\

\noindent{\LARGE Recurrent Neural Networks in the Eye of  Differential Equations}

\ \\
{\bf \large Murphy Yuezhen Niu$^{\displaystyle 1, \displaystyle 2, \displaystyle 3}$\footnote{yuezhenniu@gmail.com}, Isaac L. Chuang$^{\displaystyle 1, \displaystyle 2}$, Lior Horesh$^{\displaystyle 4,\displaystyle 5}$}\\
{$^{\displaystyle 1}$Department of Physics, Massachusetts Institute of Technology, 77 Massachusetts Avenue, Cambridge, MA, 02139}\\
{$^{\displaystyle 2}$Research Laboratory of Electronics, Massachusetts Institute of Technology, 77 Massachusetts Avenue, Cambridge, MA, 02139}\\
{$^{\displaystyle 3}$Google Inc., 340 Main Street, Venice, CA 90291}\\
{$^{\displaystyle 4}$Mathematics of AI, IBM Research, Yorktown Height, NY 10598}\\
{$^{\displaystyle 5}$MIT-IBM Watson AI Lab, Cambridge, MA 02142}\\
 
\noindent{\bf Keywords:} recurrent neural network, ordinary differential equation, Runge-Kutta method, stability analysis, temporal dynamics of Neural Networks

\thispagestyle{empty}

%
\begin{center} {\bf Abstract} \end{center}

To understand the fundamental trade-offs between   training stability, temporal dynamics and architectural complexity of  recurrent Neural Networks~(RNNs), we directly analyze RNN architectures using numerical methods of ordinary differential equations~(ODEs). We define a general family of  RNNs--the  ODERNNs--by relating the composition rules of RNNs to  integration methods of ODEs at discrete time steps.  We show that the degree of RNN's functional nonlinearity $n$ and the  range  of its temporal memory  $t$   can be mapped to the corresponding stage of Runge-Kutta recursion and the order of time-derivative of the  ODEs.   We prove that  popular RNN architectures, such as LSTM and URNN, fit into different orders of $n$-$t$-ODERNNs. This exact correspondence  between RNN and ODE    helps us to establish   the  sufficient conditions for  RNN training stability  and  facilitates more flexible top-down designs of new RNN architectures using  large varieties of toolboxes from numerical integration of ODEs.  We  provide such an example:   Quantum-inspired  Universal computing  Neural Network~(QUNN), which reduces the required number of  training parameters from  polynomial in both data length and temporal memory length to only linear in temporal memory length.
 




\section{Introduction}
Exciting progress~\citep{haber2017stable,Lu2018,Haber2018,chen2018} has been made to unveil the common nature behind  various  transformations  used in machine learning models,  such as  Neural Networks~\cite{Simon1994} and normalizing flows~\citep{Rezende2015}, realized by a sequence  of transformations between hidden states: these iterative updates can be viewed as integration of either discrete or continuous differential equations. 
Such startling correspondence not only deepens our  understanding of the  inner workings of neural network based machine learning algorithms, but also offer advanced numerical integration methods obtained over the past century to the design of better learning architectures.

Haber et. al. are the first to map the residual neural network's~(ResNet's) composition rules between hidden variables to the \textit{Euler discretization} of  continuous differential equations, and   the stability of ResNet training to the stability of the equivalent numerical integration methods. Leveraging such mapping, they   significantly improve ResNet's stability by choosing the appropriate weight matrices whose spectrum properties guarantee its stable propagation. However, their analysis is limited to one kind of numerical integration method applied to ResNet.
More recently, Chen et. al. replace the conventional neural network with its continuous limit: ordinary differential equations~(ODEs). These neural ODEs enjoy many advantages over conventional Neural Networks: back-propagation is replaced by integration of   conjugate variables representing the gradients of the hidden variables; stability is improved with the use of  adaptive numerical integration methods for ODEs; it can   learn efficiently from   time-sequential data that are generated at unevenly separated physical times; and other improvements in the parameter and memory efficiencies. Yet such fully continuous extension of neural network also faces its own challenges: it is inconvenient to use mini-batches with neural ODEs; specific error in the backward integration of conjugate variable for state trajectory reconstruction can be amplified,  and neural ODE's large scale implementation cannot directly benefit from the emerging hardware developed for tensor  multiplication. 
 
Since temporal discretization  is nonetheless unavoidable in the machine-level integration of  ODEs, why not   keep the neural network paradigm but  include a larger family of ODE integration methods in addition to  Euler discretization? A more generalized correspondence   will open the door to complex neural network architectures inspired by  physical dynamics represented by ODEs. In particular, we are interested in recurrent Neural Networks~(RNNs) for their  generality~(conventional deep Neural Networks can be regarded as the trivial type of RNN with trivial recurrence) and their capability to learn complex dynamics that require  temporal  memories.

As indispensable tools for machine translation, robotic control, speech recognition and various time-sequential   tasks,  RNNs are nonetheless limited in their application due to their susceptibility to  training instability that can be amplified by the recurrent connectivity. Various   architectural  redesigns are introduced   to mitigate this  problem~\citep{hochreiter1997long,cho2014learning,wermter1999hybrid,jaeger2007optimization,bengio2013advances,cho2014learning,koutnik2014clockwork,mhaskar2016deep,arjovsky2016unitary,jing2016tunable}. These improved stability guarantees also come  with an expense of additional architectural complexity ~\citep{jozefowicz2015empirical,karpathy2015visualizing,alpay2016learning,greff2017lstm}. They point to an  underlying trade-off between  the stability,  temporal dynamics and architectural complexity of RNN that  is yet to be found. Establishing specific connections between   more general ODE methods with composition rules of RNN architectures can be the first step towards understanding this stability-complexity trade-off.

 
In fact, \textit{Runge-Kutta methods}~\citep{Runge1895,Kutta1901} are generalizations to Euler's method, which numerically solve ODEs with higher orders of functional non-linearity through higher stages of recursion in their discrete integration rules. Since a higher order time-derivative can be transformed to coupled first order ODEs, an $n$-stage Runge-Kutta with $t$-coupled   variables thus represents a $t$th order ODE with $n$th order temporal non-linearity.  Different orders of Runge-Kutta methods  not only facilitate different orders of convergence guarantees to the numerical integration; they also provide a simple but accurate understanding of the underlying dynamics embodied by the ODEs.

In this work, we establish  critical connections between RNN and ODE: the temporal dynamics of RNN architectures can be represented by a specific numerical integration method for a set of ODEs of a given order.  Our result  elucidates a fundamental trade-off between training stability, temporal dynamics, and architectural complexity of RNN:  network's  stability and complexity of temporal dynamics can be increased by increasing the length of temporal memory~\citep{koutnik2014clockwork} and the degree of temporal non-linearity, which on the other hand demands more non-local composition rules and thus higher complexity of the network architecture as predicted by the corresponding   ODE integration method.
This insight has practical implications when applying RNNs to  real-world problems. On the one hand, additional information about the training data, such as its temporal correlation length   obtained by lower level prepossessing,  can be valuable for the choice of  RNN architectures. On the other hand, one can design unconventional RNN architectures inspired by ODE counterparts for increased ability to represent complex temporal dynamics. For example, as opposed to autonomous ODEs which do  not explicitly depend on the physical time of the incoming temporal data,   RNNs based on non-autonomous dynamical ODEs have weight matrices  specifically dependent  on the input at each iteration after the training, or more concisely are dynamical.
This captures and generalizes now-common extensions to traditional RNN structures, such as in the Neural Turing machine~\citep{graves2014neural}  which adds a write and read gated function to the data-independent network to facilitate the learning of arbitrary procedures.   Illustrating the potential of this direction, we provide here one such dynamical weight RNN construction, inspired by a quantum algorithm for realizing universal quantum computation through the preparation of the ground state of a "clock-Hamiltonian"~\citep{aharonov2008}. A clock-Hamiltonian represents the dynamical map between input and output of each temporal update, and is therefore specifically input dependent.

Using this specific correspondence between general ODE integration methods and RNN composition rules, we also identify a new property about RNN architecture: the order of non-linearity in its represented  temporal dynamics. Traditionally, the stability of RNN is considered only with respect to its memory scale--represented by the order of time-derivatives in discrete ODEs. However, we realized that the range of connectivity between hidden layers in RNN can also affect the order of temporal non-linearity reflected in different  stage of Runge-Kutta recursion. This offers us an additional insight to the inner workings of RNNs that is helpful for designing more suitable architectures given partial knowledge of the underlying physical dynamics of the data.

The structure of the paper are summarized as follows. In Sec.~\ref{ResRNN} we identify  RNN's temporal memory scale $t$ and the  degree of  non-linearity   $n$ to its underlying architecture  through an explicit correspondence to ODEs, by analyzing the   discrete  ODEs integration method  that matches the  RNN composition rules between hidden variables. We show that
$t$ and $n$ are respectively determined by the order of time-derivative and  the order of non-linearity of the ODEs that represent the RNN's dynamics.  We also provide sufficient condition for training stability for any $n$-$t$-ODERNN.
 Existing RNN architectures can thus be comprehended on the same ground according to their memory scale and the  non-linearity of their dynamics~(Table.~\ref{Table0}). In Sec.~\ref{UCNNSEC}  we  provide an example of constructing new  RNN architectures by   choosing the   appropriate underlying ODE dynamics first: Quantum inspired Universal computing recurrent Neural Network~(QUNN). QUNN is unconventional in its specific time-dependence in the weight matrix construction. The number of training parameters in QUNN  grows linearly with the temporal correlation length between input data which is otherwise independent of the dimension of  data itself. We discuss the implication of our results in Sec.~\ref{conclude}.

\begin{table} 
\centering
\scalebox{0.8}{
\begin{tabular}{c @{\hspace{0.2cm}}  c @{\hspace{0.2cm}} c @{\hspace{0.2cm}} c @{\hspace{0.3cm}}  c@{\hspace{0.3cm}} c} \hline \hline   LSTM & GRU & URNN & CW-RNN & QUNN  \\\hline
  $2$-$L$-ODERNN   &  $2$-$L$-ODERNN & $2$-$L$-ODERNN & $L$-$L$-ODERNN& $L$-$L$-ODERNN\\
\hline \hline
\end{tabular} }
\caption{Categorization of  RNN architectures according to their temporal memory scale and their order of non-linearity. LSTM~\citep{hochreiter1997long}: long-term short-term memory RNN with $L$ hidden layers. GRU~\citep{cho2014learning}: gate model recurrent neural network with $L$ hidden layers. URNN~\citep{arjovsky2016unitary}: unitary evolution recurrent neural network with $L$ hidden layers. CW-RNN~\citep{koutnik2014clockwork}: clockwork recurrent neural network with $L$ hidden layers. QUNN: quantum universal computing recurrent neural network with $L$ hidden layers. $n$-$t$-ODERNN: recurrent neural network whose temporal dynamics is represented by   a discrete integration of $t^{th}$ order ODE recursion relation using $n^{th}$ order recursion methods. } \label{Table0}
\end{table}

 \section{Stable Recurrent Neural Network}\label{ResRNN}
 
%
%
%
The success of supervised machine learning techniques  depends on the stability, the representability and the computational overhead associated with  the proposed training architecture and training data. Careful engineering of  network's architectures are necessary to realize these desirable properties since a generic neural network without any structure  is susceptible to exploding or vanishing gradients~\citep{bengio1994learning,pascanu2013difficulty}.  

The groundbreaking work by Haber et. al. \citep{haber2017stable} provides an elegant solution to  guaranteeing the stability of deep Neural Networks: understand the  neural network forward and backward propagation as a form of integration of discrete ODEs.
As an example, we look at a type of  ResNet proposed in~\citep{haber2017stable}.  Let   $l^{th}$ layer  of hidden variable be   $Y_l \in \mathcal{R}^{s\times p} $ and bias be  $b_l \in \mathcal{R}^{s\times p} $, to ensure the stability of propagation. They introduce a conjugate variable $Z_{l \pm \frac{1}{2}} \in \mathcal{R}^{s\times p} $ as a intermediate step such that the propagation of neural network is described  by 
\begin{align}\label{LEAPfrog}
&Z_{l+\frac{1}{2}}= Z_{l-\frac{1}{2}} - h_l\sigma(W_l^{\top} Y_l +  b_l), \,\, Y_{l+1} = Y_l+ \sigma(W_lZ_{l+\frac{1}{2}} + b_l).
\end{align}
The dynamics of the above discrete ODE is  stable   regardless of the form of weight matrix  $W_l$~\citep{haber2017stable}. 
 


If RNN can be trained to represent temporal structures of physical data,  its stability and complexity should also be understandable through the physical dynamics represented by ODEs.  
We generalized the method  by~\citet{haber2017stable}  introduced above to include higher order non-linearity and higher order time-derivative ODEs and to apply it larger family of neural network that include   existing architectures of RNNs as special cases.  We   define  an ODE recurrent neural network with $n^{th}$  order  in non-linearity and $t^{th}$ order in time-derivative ~($n$-$t$-ODERNN) according to its  propagation rule: the update of $n$-$t$-ODERNN can be mapped to   a generalized $n$-stage Runge–Kutta integration with $t$ coupled variables.  The specific choice of Runge-Kutta method is not essential to such generalization, and can be replaced by other integration method that provides different architecture ansatz. Such generalization help us to provide a sufficient condition for the stability criteria for any $n$-$t$-ODERNN. We then categorize several existing RNN architectures into different  $n$-$t$-ODERNN according to their temporal memory scale and degree of non-linearity. Lastly, we  define the $n$-$2$-ODERNN with anti-Hermitian weight matrices as $n$-ARNN and prove the stability of  1-ARNN and 2-ARNN.

%
%
%
%

\subsection{$n^{th}$ order ODE Recurrent Neural Network }
 
%

\noindent\textbf{Definition 1. } An  ODE recurrent neural network of $n^{th}$  order in non-linearity $t^{th}$ order in gradient~($n$-$t$-ODERNN), with integers $n, t\geq 1$ and    $k\in[n]$, is described  by the update rule between   input state value $Y_{l} \in \mathcal{R}^{s} $ at time step $l$,  the hidden variables of the $k^{th}$ layer as $K_{k}\in \mathcal{R}^{p} $ with $1 \leq k  \leq n$,  and output state value $Y_{l+1} \in \mathcal{R}^{s} $ as
\begin{align}\label{NTODERNN1}
& K_{ 1 }=  \sigma_{1 } \left( W_{1}   Y_{ l } + b_{1}  \right),\\
&K_{ q } =\gamma_{q}  K_{   q-1   }+   \kappa_{q}  \sigma_{q }\left( W_{q } Y_{ l-  t _{q-1}  } + b_{q} +h  \sum_{k=1}^{n}\alpha_{q,k}  K_{ k } \right),\\\label{NTODERNN2}
&Y_{l + 1} = \gamma_{ n+1 }  Y_{ l } + \kappa_{ n+1 }   \sigma_{q+1 } \left( W_{  n+1  } Y_{ l   -  t _{n-1} } + b_{  n+1  } + h  \sum_{k=1}^{n}\beta_{k} K_{   k  }  \right)
\end{align}
where $  2\leq q \leq n$;   the   time corresponding to each hidden layer  obeys $  t_k =  \lfloor t \frac{ k}{n}\rfloor$ with the number of inputs in time coupled by the $n$ hidden layers being $t$;   the point-wise activation function  $\sigma_{*}(\circ): \mathcal{R}^{p} \to \mathcal{R}^{p} $   at each layer is a nonlinear map that preserves  the dimension of the input;   the weight matrix at each layer is represented by $W_{q} \in  \mathcal{R}^{q_2\times q_1}$ where $q_1$ is the dimension of the input variable and $q_2$ is the dimension of the output variable of that layer; the scalar constant $h$ changes the rate of update of hidden variables between layers, the vector $b_q$ is the bias for the $q$th hidden layer, $\beta_k, \gamma_{q}, \kappa_{q}$ and $\alpha_{q,k} \in \mathcal{R}^{p\times p}$ are   matrices served to rescale and rotate the hidden variables.

To facilitate later discussions based on Definition 1, we name $\alpha_{q,k}$ the  Runge–Kutta matrix and   $\{\beta_k\}$ the Runge-Kutta weight matrices  following similar notation  in the  numerical integration of ODEs. Lastly, we define Burrage-Butcher tensor $Q$ named after its first inventors with each element defined by
 \begin{align}
        Q_{i,j}= \beta_i \alpha_{ij} + \beta_j \alpha_{ji}- \beta_i \beta_j^{\top}
    \end{align}
    which will be an important quantity in the stability analysis of the represented dynamics to be discussed in Theorem 1.

\begin{figure}[h]
\begin{center}
\includegraphics[width=0.5\linewidth]{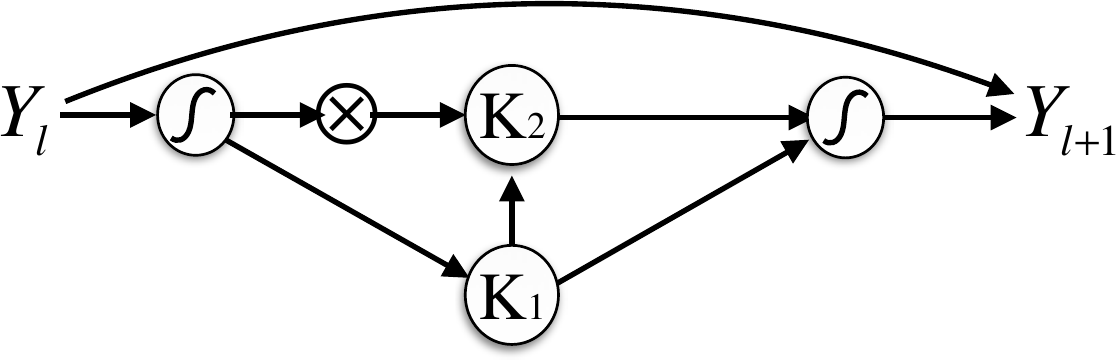}
\caption{Diagrammatic representation of $2$-$2$-ODERNN, where the $\int$ sign inside a circle represents the nonlinear activation function, and the $\otimes$ represents the time-delayed feed forward. Each arrow represent the multiplication by a re-scaling factor: $\gamma_3$ for the arrow from $Y_l$ to $Y_{l+1}$, $\kappa_2$ for the arrow from $\int$ sign to $K_1$ , $\gamma_2$ for the arrow from $K_1$ to $K_2$, and $\beta_k$ for the arrow from $K_k$ to $\int$ sign.
\label{22ODERNN}}
\end{center}
\end{figure}

\begin{figure}[h]
\begin{center}
\includegraphics[width=0.6\linewidth]{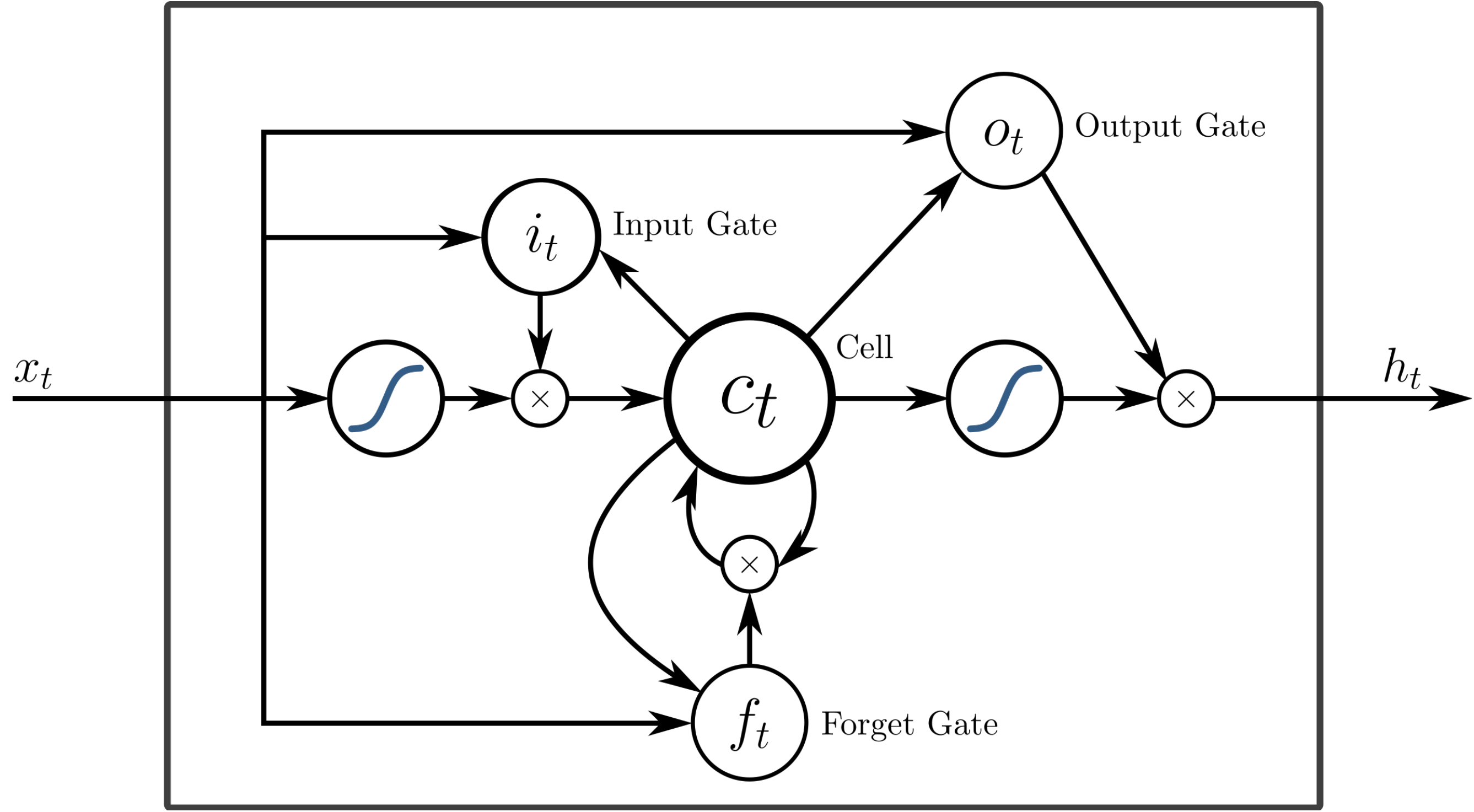}\singlespacing
\caption{ Diagrammatic representation of peehole LSTM taken from \cite{WikiLSTM}, where the $\int$ sign inside a circle represents the nonlinear activation function, and the $\otimes$ represents the time-delayed feed forward.
\label{LSTMdiagramnew}}
\end{center}
\end{figure}

\noindent\textbf{Example $2$-$2$-ODERNN }. A $2$-$2$-ODERNN is a four layer RNN with two hidden layers that obey the composition rules:
\begin{align}\label{NTODERNN1}
& K_{ 1 }=  \sigma_{1 } \left( W_{1}   Y_{ l } + b_{1}  \right),\\
&K_{ 2 } =\gamma_{2}  K_{ 1 }+   \kappa_{2}  \sigma_{2}\left( W_{2} Y_{ l -  1  } + b_{2} +h   \alpha_{2,1}  K_{1} \right),\\\label{NTODERNN2}
&Y_{l + 1} = \gamma_{ 3}  Y_{ l } + \kappa_{ 3 }   \sigma_{3 } \left( W_{ 3 } Y_{ l  -1} + b_{3 } + h  \sum_{k=1}^{2}\beta_{k} K_{   k  }  \right)
\end{align}
which is illustrated in Fig.~\ref{22ODERNN}. In comparison the connectivity for LSTM is represented in Fig.~\ref{LSTMdiagramnew}, where the connection between input and output in Fig.~\ref{22ODERNN} is replaced by the output gate $o_t$, the first hidden layer $K_1$ is replaced by forget gate, and the second hidden layer $K_2$ is replaced by input gate and memory cell.


Definition 1 can be compared with the explicit or implicit Runge-Kutta method for solving $\frac{d}{dt}\vec{y}=f(\vec{y}, t)$ where the vector $\vec{y}$ is an unknown function of time that ODE solution provides. Recall that a $t$th order ODE can be mapped to the first order ODE with  $\vec{y}$ of length $t$, each element of which is proportional to different order of discrete time derivative of the original variable. The   $n$-stage explicit or implicit Rugge-Kutta method can be generalized to the following form for the solution to the ODE at the discrete time   $t_k$ with time step $t_k-t_{k-1}= \delta$ given the solution  $\vec{y}_{k-1}$ at the previous time step through the following iteration:
\begin{align}
&\vec{y}_k = \vec{y}_{k-1} + \delta \sum_{i=1}^n  e_i\vec{k}_i\\
&\vec{k}_q = f(d_q \vec{y}_{k-1} + \sum_{j=1}^{n } \delta a_{qj}\vec{k}_j, t_{k-1}+c_q \delta)  
\end{align}
where $ a_{q,j}, e_i, c_q, d_q$ are square matrices and determine the corresponding integration method. For example, if we set $a_{q,j}=0$ for all $q \leq j$, it gives us an explicit Runge-Kutta method, otherwise it corresponds to an implicit Runge-Kutta method. The difference between the two methods is in the additional requirement of solving the linear dependence of $\{\vec{k}_q \}$ in each iteration of implicit method, which   lower the requirements on $f()$ for the numerical stability of the integration. If we treat the $k$th hidden layer from the ODERNN as the $k$th stage of integration method above, and choose the matrix $d_k$ to pick out the $k$th derivative which in discrete time-step corresponds to the variable separated by $k\delta$, the  order of time derivative and the order of functional non-linearity of $n$-$t$-ODERNN becomes self-evident.  

With this explicit connection, we can directly apply the stability analysis of Runge-Kutta method to the ODERNN  with  Theorem 1, which utilizes the notion of BN-stability specified below in Definition 2. BN-stability  was first proposed by  \citep{Dahlquist1979} to investigate stability of numerical schemes applied to nonlinear systems  satisfying a monotonicity condition, which is a generalization of the ``A" stability   for linear systems and   widely used in analyzing the stability of high order Runge-Kutta methods.

\noindent\textbf{Definition 2.}The integration method for solving the nonlinear   discretized ODE system of equations $\dot{\vec{y}} = f(\vec{y},t)$ is  BN-stable if it   satisfies the following requirements. It is monotonic: the inner product between the variable vector and the function vector is non-negative $\langle f(\vec{x},t)-f(\vec{y},t), \vec{x}-\vec{y}\rangle \leq 0$ for $t\geq 0; \vec{x}, \vec{y}\in \mathcal{R}^s$; and a small perturbation at the initial state   $\vec{y^\prime}_0=\vec{y}_0+\delta_0$ does not amplify as step size increases: for any $k$ 
\begin{align}
    ||\vec{y}_{k-1}-\vec{y^\prime}_{k-1}|| \leq    ||\vec{y}_k-\vec{y^\prime}_k||.
\end{align}
\vspace{5pt}

Based on Definition 2, we are ready to provide a stability guarantee for the ODERNNs in the following theorem.
 
\noindent\textbf{Theorem 1.} An $n$-$t$-ODERNN given by Eq.~(\ref{NTODERNN1}) is BN-stable if it satisfies the following conditions:
\begin{enumerate}
    \item[I.] The Burrage-Butcher tensor $Q$
    is positive semi-definite.
    
    \item[II.] For any $k\in [n]$, the matrix $\beta_k$ is positive semi-definite.
\end{enumerate}

\noindent\textit{Proof}: Since the composition rule of $n$-$t$-ODERNN can be mapped to that of an $n$ stage general implicit Runge-Kutta method, the BN-stability proof for Theorem 1.4 from \cite{Spijker1980} directly applies.  The monotonicity requirement~\citep{Dahlquist1979}  in Definition 2 is not sensitive to the gradient of the function and can also be replaced by $\langle f(\vec{x},t)-f(\vec{y},t)$, where we use $\vec{x}$ and $\vec{y} $ to represent the input and output of each hidden layer of RNN, which obeys $\langle \vec{x}-\vec{y}\rangle \geq 0$ for $t\geq 0; \vec{x}, \vec{y}\in \mathcal{R}^s$, which is naturally satisfied by the rectified  linear function or tanh. 

\subsection{Categorization of Existing RNNs}
We apply this ODERNN framework to analyze some of the most widely used RNN architectures in regard to their non-linearity and memory scale of their underlying dynamics.

\begin{table} 
\centering 
\scalebox{0.8}{
\begin{tabular}{c @{\hspace{0.2cm}}  c @{\hspace{0.2cm}} c @{\hspace{0.2cm}}c @{\hspace{0.2cm}} c} \hline \hline   & traditional RNN (LSTM)&\vline  &physical RNN (ODERNN) \\\hline
$Y_l$&  input at time step $l$&\vline&  state variable  at time step $l$  \\\hline 
$K_{l_j}$ & $j^{th}$  hidden layer &\vline&  $ j^{th}$ order increment of the gradient slope  \\
 $\gamma_{l_j}$& forget gate activation  &\vline& energy dissipation rate \\\hline
 $\alpha_{i,j}$ &weight matrix for hidden variable &\vline & weight of $i^{th}$ increment in $j^{th}$ order slope\\\hline
 $\kappa_{l_j}$&input gate's activation &\vline & re-scale factor of normalized gradient function \\\hline
  $\sigma_{l_j}$ &activation function of $j^{th}$ hidden layer &\vline & gradient function \\
\hline \hline
\end{tabular} }
\caption{Comparison of the LSTM architecture and $n^{th}$ order ODERNN structure. } \label{Table2}
\end{table}

\noindent\textbf{Definition 3.}  A LSTM with $L$ non-trivial hidden layers $\{K^l_t\},$ with $ l \in [L]$, at time step $t$ obeys the following propagation rules between hidden variables. Each  $K^l_t$ is of dimension $n$ and is updated at each time step as follows:
\begin{align}
   & \left(\begin{matrix}
    i\\
    f\\
    o\\
    g
    \end{matrix}\right) = \left(\begin{matrix}
    \sigma_1\\
    \sigma_2\\
    \sigma_3\\
    \sigma_4
    \end{matrix}\right)W^{l}\left(\begin{matrix}
    K_{t}^{l-1}\\
    K_{t-1}^l
    \end{matrix}\right)\\\label{lstm2}
   &  c_t^l = f \circ c_{t-1}^l + i \circ g\\\label{lstm3}
  &  K_{t}^l= o \circ \tanh(c_t^l)
    \end{align}
where $W^l$ is of dimension $4n\times 2n$, $\circ$ represents the element-wise product, and the four vectors, $i, f, o, g$ each of dimension $n$, with the first three controlling which  input will be saved to influence the future output according to the above update rules in Eq.~(\ref{lstm2}) and (\ref{lstm3}).

\noindent\textbf{Claim 1.} For any   LSTM with $L$ hidden layers in Definition 3, there   exists a 2-$L$-ODERNN that realizes the same input-output relation.

\textit{Proof}: For one layer RNN, we have the update rule for LSTM~\citep{hochreiter1997long} as: 
\begin{align}
&K_t= f_t  \circ K_{t-1} + i_t \circ \sigma_2( W_c Y_{t-1} + U_c K_{t-1} + b_c),\\
&Y_t= o_t \circ \sigma_1(K_t)
\end{align}  with vector coefficient determined by
\begin{align}
&f_t=\sigma(W_f Y_{t-1} + U_f K_{t-1} + b_f),\\
&i_t=\sigma(W_i Y_{t-1} + U_i K_{t-1} + b_i),\,o_t=\sigma(W_o Y_{t-1} + U_o K_{t-1} + b_o)
\end{align}
which is equivalent to setting $n=2$, $t=1$, $\gamma_{ 2 }=D[f_t], \kappa_{ 2}=D[i_t], W_{ 2  }=W_c,   b_{ 2 } =b_c, h\alpha_{21}=U_c$ and $\gamma_{ 2}=0, \kappa_{2}=o_t $ in $n$-$t$-ODERNN.  Notice that the weight matrix $W_q$ in ODERNN can depend on time and is therefore able to include the memory dependency from $K_{t-1}$.
We use $D[a]$ to represent a $p\times p$ diagonal matrix with each diagonal element equal to each element of the vector $a$ of length $p$. This is because the element-wise product between two vectors can be re-written as diagonal matrix matrix multiplication with the second vector: $a \circ b= D[a]b$. 

For multi-layer LSTM with $L$ hidden layers, the only change is that the diagonal matrices $D[f_t], D[i_t]$  and $ D[o_t] $ are generalized to $D[f_t^l],  D[i_t^l] $  and $D[o_t^l] $, which   not only depend  on the hidden variable of the same layer from the previous time step, but also the hidden variable of the same time step from a previous layer:
\begin{align}
&f_t^l=\sigma(W_f K_{t}^{l-1} + U_f K_{t-1}^l + b_f^l), \\
&i_t^l=\sigma(W_i K_{t}^{l-1} + U_i K_{t-1}^l + b_i^l),\\
&o_t^l=\sigma(W_o K_{t}^{l-1} + U_o K_{t-1}^l + b_o^l)
\end{align}
where $K_{t}^{0}=Y_{t-1}$, and thus the non-linearity of the ODE increases by one when the number of hidden layers increases by one, thus gives  $L$-2-ODERNN for a $L$ layer architecture, Q. E. D..

\noindent\textbf{Definition 4.}  A Gated Recurrent Unit~(GRU)  with $L$ non-trivial hidden layers $\{K^l_t\},$ with $ l \in [L]$, at time step $t$ obey the following propagation rules between hidden variables. Each  $K^l_t$ is of dimension $n$ and is updated at each time step according to:
\begin{align}
    & \left(\begin{matrix}
    r\\
    z
    \end{matrix}\right) = \left(\begin{matrix}
    \sigma\\
    \sigma 
    \end{matrix}\right)W^{l}\left(\begin{matrix}
    K_{t}^{l-1}\\
    K_{t-1}^l
    \end{matrix}\right)\\\label{GRU2} 
  &  K_{t}^l= (1-z)   \circ K_{t-1}^l + z\circ \tanh(W_x^l K_{t}^{l-1} + W_g^l r\circ K_{t-1}^l)
\end{align}
where weight matrix $W^l$ is of dimension $2n\times 2n$, and weight matrices $W_x^l$ and $W_g^l$ are both of dimension $n\times n$, $\sigma $ represents a given point-wise non-linearity.

\noindent\textbf{Claim 2.} For any GRU  with $L$ hidden layers as in Definition 4, there exists a 2-$L$-ODERNN that realizes the same input-output relation between each layer of hidden variables. 

For one layer GRU~\citep{cho2014learning}, we have the update rule as:
\begin{align}
&Y_t= (1-z)\circ Y_{t-1} + z\circ \tanh\left( W_t Y_{t-1} +W_g r\circ Y_{t-1} \right),\\
&\text{with} \,\, z=\sigma(W_l^z  Y_{t-1}), \,\, r=\sigma(W_l^r  Y_{t-1})
\end{align}
we can rewrite $r\circ Y_{t-1} $ as $\sigma^\prime(W_l^qY_{t-1} )$, rewrite $\sigma(W_l^z Y)\circ \tanh( W_t Y)$ as $\sigma^{''}(W_{t,l}Y)$ and thus simplify the update rule to
\begin{align}
&Y_t= (1-z)\circ Y_{t-1} +  \sigma^{''}\left( W_{t,l} Y_{t-1} + W_{t,l}^{-1} W_g\sigma^\prime(W_l^qY_{t-1} )\right)
\end{align}
which is equivalent to setting  $n=t=2$, $\gamma_{2 }=D[(1-z)], \kappa_{2}=1, W_{2 }=W_{t,l},   b_{2} =0, h\alpha_{1}= W_{t,l}^{-1} W_g$   in $n$-$t$-ODERNN. This can be similarly generalized to multi-layer GRU with $L$ total hidden layers by defining the weight matrices $D[(1-z)]$ and $D[r]$ with: 
\begin{align}
&z=\sigma(W_l^z  Y_{t-1}^l+W_l^{z^\prime}  Y_{t}^{l-1}), \,\,\, r=\sigma(W_l^r  Y_{t-1}^l+W_l^{r^\prime}  Y_{t}^{l-1})
\end{align}
which for $l^{th}$ layer it corresponds to $L$-2-ODERNN. Q. E. D..

%
%

\noindent\textbf{Definition 5.}  A Unitary evolution Recurrent Neural Network~(URNN)  with $L$ non-trivial hidden layers $\{K^l_t\},$ with $ l \in [L]$, at time step $t$ given input data at time step $x_t$ obey the following propagation rules between hidden variables. Each  $K^l_t$ is of dimension $n$ and is updated at each time step according to:
\begin{align} 
  K_{t}^l= \sigma(  W_lK_t^{l-1} + V_l x_t)
\end{align}
where the weight matrix $W_l$ is of dimension $n\times n.$

\noindent\textbf{Claim 3.} For any  URNN~\citep{arjovsky2016unitary} in Definition 5, there exists  a $2$-$L$-ODERNN that realizes the same transformation of hidden variables.
 
\noindent\textit{Proof}: The propagation rule of URNN between the hidden variable at time step $t$ of the $l$th layer with $1 \leq l\leq T$ can be realized by a $2$-$L$-ODERNN by setting $\gamma_{q}=0$ and $W_q=0$ and choosing $n=L$ and $t=2$ in Eq.~(\ref{NTODERNN1})--(\ref{NTODERNN2}). URNN thus belongs to $2$-$L$-ODERNN. Q.E.D.

\noindent\textbf{Claim 4.}  There exists a $L$-$L$-ODERNN that realizes clockwork RNN~(\cite{koutnik2014clockwork}) with $L$  clocks.

\noindent\textit{Proof}: The propagation rule for CW-RNN between input $Y_t$ at time step $t$, hidden layers at the same time step $ K_t $ as well as from the previous time step  $ K_{t-1} $ and output $Y_{t+1}$ is  described by:
\begin{align}
&K_t=   \sigma_h\left(W_H(t) K_{t-1} +  W_I(t) Y_t  \right), \,\,
Y_{t+1}= \sigma_o\left(W_o K_t \right)
\end{align}
where the dynamical weight matrices $W_H(t)$ and $W_I(t) $ are structured to store memory of previous time steps in into different blocks with increasing duration of time delays such that effectively one can rewrite $W_H(t) K_{t-1}=\sum_{j<t-1}\left[ W_j\sigma_j^{t-j-1}W_H(j)K_j+ W_I(j)Y_j\right] $ contributions from all   previous step iteratively,  and so is the clock structure in $W_I(t) Y_t $ which contributes to all hidden layers after $t$. This   is equivalent to setting $n=t=L$ and  $\gamma_{q}=0$  in Eq.~(\ref{NTODERNN1})--(\ref{NTODERNN2}). CW-RNN thus belongs to $L$-$L$-ODERNN. Q.E.D.

\subsection{$n$-$t$-ARNN}\label{ARNNSec}
Apart from the categorization of some of existing RNN architectures,   this ODE methodology can also be applied to design new kinds of RNN  starting first by choosing   the order of the ODERNN and the   weight matrices. Now we showcase the advantage of such top-down construction of RNN architecture: its length of temporal memory scale and degree of temporal non-linearity  are determined by design. Its application in reducing the architectural complexity while guaranteeing the stability and representability will be demonstrated in the next section.

\noindent\textbf{Definition 3.} The $n^{th}$ order ODE anti-Hermitian recurrent neural network~($n$-$t$-ARNN) corresponds   $n$-$t$-ODERNN with   anti-Hermitian weight matrices.

\noindent\textbf{Theorem 2.} 1-2-ARNN with monotonic activation function  $\sigma_{*}(\cdot): \mathcal{R}^{n} \to \mathcal{R}^{n} $ and purely imaginary anti-Hermitian weight matrix   is  stable for   $h$  that satisfies $ ||h \max_{k\in n} \lambda_k[W_{1}]||<1$, where $\lambda_k[W_{1}]$ represents the $k$th eigenvalue of $W_{1}$. 

\textit{Proof}: This will be  proven in Theorem 5, where the original complex anti-Hermitian matrix is embedded into a Hilbert space twice as large such that a purely imaginary anti-Hermitian weight matrix guarantees the stability of the first order integration method.  Q.E.D.

\noindent\textbf{Definition 4}. An integration method that correspond to a map $\Phi: M\to M $ for linear space $M$ is reversible with respect to a reversible differential equation that represent a differential map $\rho$ if $\Phi$ exists and the following holds:
\begin{align}
 \rho(\Phi) = \Phi^{-1}(\rho).
\end{align}

\noindent \textbf{Theorem  3.} Both 2-2-ARNN and 1-2-ARNN are reversible.

\textit{Proof}: It is not hard to see that 2-ARNN corresponds to the first order mid-point integration  and 1-ARNN corresponds to the symplectic Euler integration. Their reversibilities are guaranteed by the reversibility of these two integration schemes inside the stble regime~\citep{haber2017stable}.  Q.E.D.

It is notable that the definition of $n$-$t$-ODERNN  does not restrict   weight matrices to be independent of the input to each recursion step. This setup is less restrictive than conventional definition of RNN and is indispensable for generalizing various architectures of RNN under the same framework. Such generalization, however, is natural to its  ODE counterparts: a generic ODE can be non-autonomous.

The unification of different RNN architectures through $n$-$t$-ODERNN paves the way for tailoring the temporal memory scale and the degree of non-linearity of RNN architecture towards the underlying data, while   reducing the complexity overhead in the learning architecture. We showcase one such application  in   RNN design in the next section.

\section{Quantum Inspired Universal Computing Neural Network}\label{UCNNSEC}
%


Despite the complex structure of LSTM,   a simple 2-$L$-ODERNN is able to represent the same type of temporal dynamics as LSTM while providing specific stability guarantees through Theorem 1. 
It is thus tempting to construct RNN  starting from choosing the appropriate ODE counterparts first.
In this section, we provide such an example of  RNN  design, QUNN, by emulating the ODEs for quantum dynamics with the RNN architectures.

 For preparation, we  bridge the gap between RNN dynamics and the quantum dynamics in its discrete realization in Sec.~\ref{qUANTUMRNN}. We illustate the application of ODE-RNN correspondence in designing better RNN architectures in Sec.~\ref{QURNNSEC}, where we define a QUNN construction  inspired by a universal computation scheme in quantum systems first conceived by Richard Feynman~\citep{feynman1986}.  

\subsection{Quantum Dynamics in the Eye of ODEs}\label{qUANTUMRNN}
The dynamics of a quantum system can be described by a first-order ordinary differential equation, namely the  Schr\"{o}dinger  equation, where the quantum state   represented by the complex vector $Y(t)$ at time $t$ obeys: 
\begin{align}\label{Shrodinger1}
\frac{d}{dt} Y(t) = -i H(t) Y(t), 
\end{align}
where $ H(t)$ is a Hermitian matrix representing the system Hamiltonian and determining the dynamical evolution  of the quantum state.  The Hamiltonian matrix is nothing more than the   gradient with respect to time in the  first order linear ODE.  Despite such fundamental linearity, the emergent phenomena in a sub-system by tracing out certain elements in the state parameter $Y(t)$  can be highly nonlinear.  

The  quantum dynamics represented by Eq.~(\ref{Shrodinger1}) can perform  universal computation as first suggested  by Richard Feynman in~\citep{feynman1986}: any Boolean function can be encoded into the reversible evolution of a quantum system under the a system Hamiltonian that contains independent Hamiltonian terms polynomial in the number of   logical gates needed to describe  the   Boolean function.  This result is captured by the Theorem 5 in Appendix A.

The universality and reversibility of quantum dynamics inspire us to   harness previous results by Kitaev and Aharonov et. al.\citep{kitaev2002,aharonov2008} to propose a RNN ansatzs resembling the  construction of temporal dynamics through the addition of a clock-Hamiltonian. There,  the time evolution of a generic quantum system can be represented by  the lowest energy eigenstate  of a clock Hamiltonian matrix defined on an enlarged system.  This clock Hamiltonian has two unique properties: it is constructed using the quantum state trajectory of the target system, for periodic system it contains only a fixed number of Hamiltonian terms proportional to the periodicity.  Based on our newly established connection between  RNN and ODE, we can construct RNN architectures whose temporal dynamics emulate  quantum evolution under the clock Hamiltonian.

\subsection{Quantum-inspired RNN}\label{QURNNSEC}
We propose an RNN ansatz called quantum-inspired universal  recurrent neural network~(QUNN).  It reduces the number training parameters   in  temporal correlation length from quadratic in traditional architectures to linear.  We achieve this by constructing   weight matrices in hidden layers from the input data based on a quantum-inspired ansatz. Such construction exploit the knowledge of   temporal structure embodied by  the data itself. In comparison, traditional RNN weights are data independent and remain  constant after the training.   QUNN is thus able to represent a larger family of dynamical maps by utilizing the input to the network for the construction of dynamical relations.

\noindent\textbf{Definition 5.} Let the  temporal  correlation length of training data be $N$. Quantum-inspired universal  recurrent neural network~(QUNN) with $N$ hidden layers is defined as: a recurrent neural network architecture that transform the input state $ Y_l $, which could be either a binary, real, or complex vector depending on the problem type, to the subsequent output   $ Y_{l+1}$ denoted by the integer index $l\in\{1,2,...., N\}$ according the the following three stages. 

In the first stage, the incoming data is boosted to a higher dimension by a   tensor product  with a clock state vector $\vec{c}_l $ that marks the relative distance in the training data sequence from the first input  within the same data block of length $L$ followed by a nonlinear function: $K_l^1 =   \sigma_1\left( Y_l \otimes \vec{c}_l \right).$
We use $\sigma_i( \cdot)$ here and henceforth to   represent  the  monotonic and continuously differentiable  point-wise nonlinear function  of the $i^{th}$ layer. 

In the second step, the output of the first layer is passed into a $L$ layer neural network with a composition rule between the $k$th hidden layer $ K_l^k$ given by
\begin{align}\label{updateRuleQunn2}
&K_l^k= S_k K^{k-1}_l   + \sigma_2\left(H_{l} K^{k-1}_l \right),\\
&H_{l} =D_l^kW_l^k\\\label{weightmatrix10}
&D_l^k = (1-p_1(l))I\otimes I^c+  p_1(l)   \left[ I \otimes \left( \vec{c}_{l-k}\vec{c}_{l}^{\top} \right) - I \otimes \left(\vec{c}_{l}\vec{c}_{l-k}^{\top} \right) \right]\\\label{weightmatrix1}
& W_l^k =(1- p_2(l))W_{l }^{k-1} + p_2(l)\left[  (Y_{l-k }  Y_{l }^{\top})  \otimes (\vec{c}_{l -k}\vec{c}_{l}^{\top}) - (Y_{l }  Y_{l-k }^{\top})  \otimes (\vec{c}_{l}\vec{c}_{l -k}^{\top})  \right] 
\end{align}
where we use $S_k$ to represent a weight matrix similar to that for ResNet~\citep{haber2017stable} in front of the output from the previous hidden layer,  the  $D_{l}$ consists of identity matrix on both the original data space $I$ and the clock space $I_c$ weighted by $ 1-p_1(l) $ and  a   re-ordering operator weighted by a real scalar coefficient $p_1\in[0,1]$. Notice that $\vec{c}_{1 }\vec{c}_{ l-1 }^T$ permutes the order of clock states which    adds noise~\citep{Lu2018} as well as correction to possibly  temporally mislabeled  training data. Such a permutation step is in product with $W_l$, which records the flexible range of memory important for the training: the scale factor   $p_2(l)\in[0,1] $     determines the length of temporal memory of past inputs.

\begin{figure}[h]
\begin{center}
\includegraphics[width=0.8\linewidth]{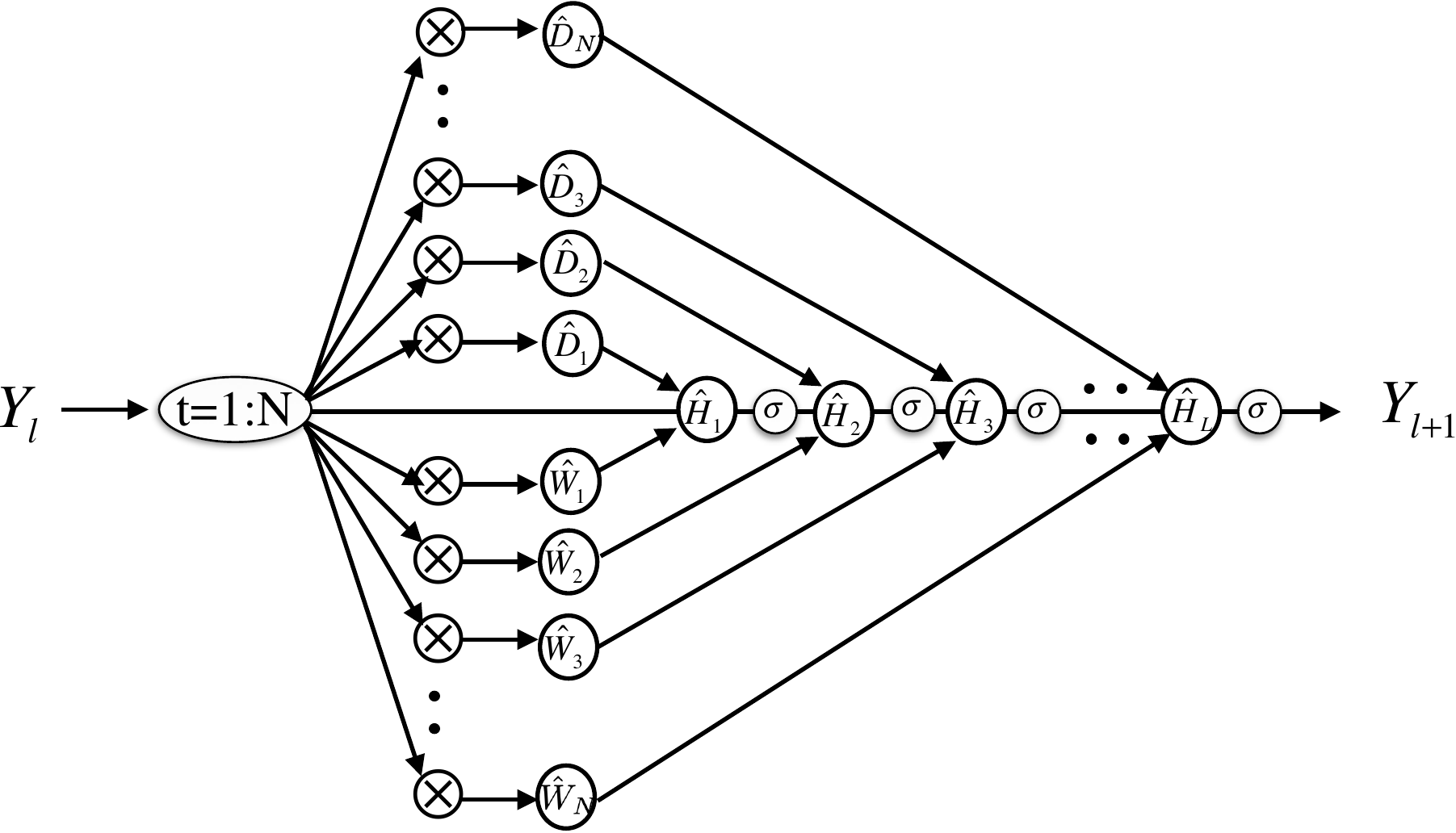}
\caption{QUNN arcitecture: $\otimes$ represents tensor product of input with  delays,  oval marked by $t= 1:N$ represents different values of delays, and $\sigma_*$ represents nonlinear activation functions at different layers.
\label{QUNNFIG1}}
\end{center}
\end{figure}

In the third step, the state is mapped back to the original data dimension by projecting it onto the corresponding clock vector for the next sequence of the incoming data string:  
\begin{align}\label{Unembedd}
 Y_{l+1} =\sigma_{L+1}\left( \text{Tr}\left[\left( I \otimes(\vec{c}_{l+1}\vec{c}_{l+1}^{\top}) \right) K_l^L\right] \right).
\end{align} 
Finally reset $l=1$ if $l\geq N$ since the temporal correlation only exists within the block of size $N$; see Fig.~\ref{QUNNFIG1} for the illustration of QUNN architecture.

Notice that the QUNN weight matrix   changes    dynamically according to Eq.~(\ref{weightmatrix1}), meaning that they evolve with the input data within the active  block of length $N$.  This is different from conventional definition of RNN where the weight matrix does not explicitly depend on the data, but such memory dependence is indirectly actuated through the gate construction such as the forgetting unit $f_i$ in LSTM~\citep{hochreiter1997long}.
The multi-layer construction of LSTM facilitates longer temporal memory as the  hidden layer number $N$ increases. The depth of the hidden layers in QUNN can also 
be interpreted  as both the order of the corresponding ODE recursion relation and the order of nonlinearity that corresponds to the stage of Runge-Kutta method, as illustrated in Fig.~\ref{QUNNFIG1}. A QUNN with $L$ hidden layers thus corresponds to a $L$-$L$-ODERNN. The  the general QUNN stability can be ensured by choosing parameters in Eq.~(\ref{updateRuleQunn2}) and (\ref{weightmatrix1}) according to the requirements in Theorem 1.
 
\begin{table} 
\centering
\begin{tabular}{c @{\hspace{0.1cm}}c @{\hspace{0.2cm}} c @{\hspace{0.2cm}}   c @{\hspace{0.2cm}}  c @{\hspace{0.2cm}} c @{\hspace{0.2cm}} c} \hline \hline &\vline& QUNN  & LSTM &URNN & $n$-$t$-ODERNN  \\\hline
memory scale&\vline &  $n$&  $2$ &  $2$ &$t$  \\\hline 
order of nonlinearity&\vline &   $n$  & $n$ & $n$  &  $n$  \\\hline
stability&\vline& Yes  & $?$ & Yes & $?$ \\\hline
depth& \vline& $n +2$   & $n$ & $n$   & $n$ \\\hline
training parameters& \vline& $O(n)$   & $O(n^2)$ & $O(n^2)$   & ? \\\hline
origin &\vline & Schr\"{o}dinger equation & Ad Hoc & Unitarity & ODE    \\
\hline \hline
\end{tabular} 
\caption{A top-down comparison between QUNN,  LSTM,  URNN and $n$-$t$-ODERNN structure.} \label{Table2} 
\end{table}

Our  top-down design of QUNN is compared with some of existing RNN architectures in Table.~\ref{Table2}.   QUNN have longer range of memory scale than both LSTM and URNN, i.e., the order of time derivatives in the corresponding ODE is higher in QUNN. This comes with a price of additional layers of embedding in RNN architecture seen in the depth difference. The data-dependent construction of QUNN weight matrices can also slow down the convergence of the    training process. And certain pre-processing of data, such as calculating the autocorrelation function, is needed to give a good estimate of $N$ to make QUNN effective.  But QUNN can reduce the total number of training paramters from LSTM, offering potential advantage to its training efficiency. This show cases the distinction between an ad hoc heuristic approach and physically inspired approach to  RNN designs.

\section{Conclusion}\label{conclude}
 
We propose an ODE theoretical framework for understanding RNN architectures's  order of non-linearity,   length of memory scales and training stability. We apply this analysis to many existing RNN architectures,   including LSTM, GRU, URNN, CW-RNN and identify the improved nonlinearity obtained by CW-RNN. Examing RNN through the eyes of ODEs help us to  design new RNN architectures inspired by dynamics and stability of different ODEs and associated integration methods.
As an example, we showcase the design of an RNN based on the ODEs of quantum dynamics of universal quantum computation.   
We  show that in the case when the temporal correlation is known and the input data comes in active blocks, QUNN provides a quadratic reduction in the number of training parameters as a function of temporal correlation length than generic LSTM architectures.
Our findings point to an exciting direction   of  developing new machine learning tools by harnessing physical  knowledge of both the  data and the neural network itself.  

A parallel   and recently published work  on designing stable RNN based on ODEs with antisymmetric weight matrices by   \cite{Chang2019} came to our attention after the completion of this work. There, a stable RNN architecture is proposed and implemented, which have important   practical applications in improving the state-of-the-art RNN performance. In comparison, we have focused on theoretical analysis on general RNN architectures, which include the RNN design from  \cite{Chang2019}  as  a specific case  of $n$-$t$-ARNN defined in Sec.\ref{ARNNSec} by setting $n=1$ and $t$ equals the number of connected hidden layers.  And our    stability proof is applicable for any RNN designed based on ODE integration methods. We aim at establishing a firm theoretical ground for a wider application of numerical toolbox in the study of Neural Networks instead of providing a ready-to-use  RNN architecture. But more heuristic testing remains to be done to fully understand the practical use of tailoring temporal non-linearity of RNNs defined in this work.

\begin{appendix}
\begin{table} 
\centering
\scalebox{0.8}{
\begin{tabular}{c @{\hspace{0.2cm}}  c @{\hspace{0.2cm}} c @{\hspace{0.2cm}} c @{\hspace{0.3cm}} c@{\hspace{0.3cm}} c@{\hspace{0.3cm}} c} \hline \hline   & column vector & row vector & matrix & inner product & tensor product & Hadamard product\\\hline
  classical& $Y_l$& $Y_l^{\top}$ &   $W_l$ & $Z_l^{\top} Y_l   $ &$Y_l\otimes Z_l $ & $Y_l\circ  Z_l $ \\\hline 
quantum & $\ket{Y_l}$   & $\bra{Y_l}$ & $\hat{W}_l$ &$ \bra{Z_l}Y_l\rangle $&$ \ket{Y_l}\otimes\ket{Z_l} $& $\hat{D}[Y_l] \ket{Z_l}$\\
\hline \hline
\end{tabular} }
\caption{Comparison of the representation of linear algebra in quantum and in classical literature.} \label{Table1}
\end{table}\vspace{-5pt}
\section{Proof of Theorem 5. }

\noindent\textbf{Theorem 5.} Any  Boolean function from the uniform family  $f: \{ 0, 1\}^n \to  \{ 0, 1 \}^n$   can be mapped to  a unique fixed point of  ODE evolution with its characteristic function containing polynomial in $n$ many parameters. 

\textit{Proof.} By definition, any polynomial-time uniform  Boolean function can be represented by a deterministic  Turing machine that runs in polynomial time $L_c$ and outputs the circuit description $\mathcal{C}_n$ if given $1^n$ as input. We only need to show that there exists a one-on-one mapping between a deterministic Turing machine of polynomial runtime and a set of ODEs that represents quantum dynamical evolution. The read and write process of a Turing machine  can be mapped a rotation in the Hilbert  space of input  and output  $\vec{r_j}, \vec{w}_j \in \mathcal{H}$ at the $j$th time step as:   
\begin{align}
U_j= \vec{w_j} \times \vec{r}_j^{\top},
\end{align}
where our use of $\ket{w_j}$ and the associated quantum notation henceforth is explained in Table~\ref{Table1}.
To keep track of the  update in its relative location within each active block, we tensor product such matrix with a time step update matrix $\ket{ j+1 }\bra{ j}^c$ which increase the book-keeping of time $j$ by one. So the overall unitary takes the form:
\begin{align}
U_j\otimes \ket{ j+1 }\bra{ j }^c = \ket{w_j}\bra{r_j}\otimes\ket{ j+1 }\bra{ j  }^c.
\end{align}
Moreover, we  add the inverse process of such Turing computing step to satisfy the Hermiticity of the overall  matrix $\hat{H}_j$: 
\begin{align}
\hat{H}_j = \ket{w_j}\bra{r_j}\otimes\ket{ j+1}\bra{ j  }^c + \ket{r_j}\bra{w_j}\otimes\ket{ j  }\bra{ j+1 }^c,
\end{align} whose eigenvalues are purely imaginary. 
This consists of one step of reversible formulation of any TM computation step. Summing up all the corresponding steps for a TM that halts at step $L_c$, we obtain a Hermitian matrix that encode both the forward and backward computation process of a conventional TM. The ground state energy  of this Hamiltonian, or the absolute value of the lowest eigenvalues,  is not necessarily zero. If we like to ensure that the ground state which corresponds to the process of TM computation step is the fixed-point of the dynamical evolution, we need to add two additional term to obtain the overall Hamiltonian:
\begin{align}\nonumber
\hat{H}_{TM}= &   \sum_{j=1}^{L_c} \left( I \otimes\ket{ j   }\bra{ j   }^c  -\ket{w_j}\bra{r_j}\otimes\ket{ j+1 }\bra{ j }^c \right.   \left. - \ket{r_j}\bra{w_j}\otimes\ket{ j  }\bra{ j+1  }^c + I \times\ket{ j+1 }\bra{ j+1 }^c \right)  
\end{align}
whose lowest eigenvalue is exactly zero~\cite{aharonov2008}.

It is   shown  in \cite{aharonov2008}  that the zero energy ground state  of the above Hamiltonian is
\begin{align}
    \ket{\psi_0}=\frac{1}{\sqrt{L_c}} \sum_{j=1}^{L_c} \ket{w_j}\otimes\ket{j+1}
\end{align}
which  is not only unique but  also separated from other eigenstates in eigenvalues by a gap. Projecting the  this ground state onto the clock state $\ket{L_c}^c$   gives   the output of the Turing machine that describe the desired Boolean function.  Since, a polynomial runtime Turing machine can be described by polynomial sized computational tap $L_c$,   the size of $U_j$ is also polynomial in the number of bits. Since the total number of terms inside this Hamiltonian equals the number of time steps $L_c$ in the deterministic Turing machine of the uniform Boolean functions,  it is also polynomial in number of bits of the Boolean function input.  
Q. E. D.

\end{appendix}
 
\bibliographystyle{iclr2019_conference}

\end{document}